\documentclass[conference,letterpaper]{IEEEtran}
\IEEEoverridecommandlockouts
\usepackage{cite}
\usepackage{amsmath,amssymb}
\usepackage{graphicx}
\usepackage{booktabs}
\usepackage{array}
\usepackage{xcolor}
\usepackage{pifont}
\usepackage{tikz}
\usetikzlibrary{arrows.meta,positioning,calc,shapes.callouts,shapes.geometric}
\usepackage[hidelinks]{hyperref}

\graphicspath{{figs/}}
\newcommand{\cmark}{\ding{51}}
\newcommand{\xmark}{--}

\begin{document}
\bstctlcite{IEEEexample:BSTcontrol}

\title{Vision-Language Truck Driving Advisories Grounded in Provincial Rules and Forecast Risks}
\author{Xintong Jiang$^{1}$, Shucheng Huang$^{2}$, Qiaoxuan Zhang$^{1}$, Jincheng Xu$^{1}$, Yukun Lu$^{1,2*}$
    \thanks{$^{1}$Dept. of Mechanical Engineering, University of New Brunswick, Fredericton, NB E3B 5A3, Canada}%
    \thanks{$^{2}$Dept. of Mechanical and Mechatronics Engineering, University of Waterloo, Waterloo, ON N2L 3G1, Canada}%
    \thanks{*Corresponding author: Yukun Lu (email: yukun.lu@unb.ca).}    
}

\maketitle


\begin{abstract}
Vision-language models (VLMs) can generate context-aware driving advisories, but their outputs may conflict with road conditions or traffic rules, and they typically address a hazard only once it is visible. This matters for long-haul trucking, where heavy trucks need long stopping distances and a single trip may cross several jurisdictions with different traffic regulations. This paper proposes a data generation and distillation framework that integrates structured scene understanding, risk forecasting, and traffic-rule retrieval to generate truck driving decisions and natural-language advisories, with a consistency check on every record. The records, written by an 8B teacher, are then used to distill a 3B-parameter student model for onboard deployment. We also introduce a dashcam dataset covering four Canadian provinces, comprising 11,142 annotated lane-marking frames and 9,494 driving segments. On held-out driving data across all four provinces, the student passes the consistency check in 94.6\% of the segments, against 20.2\% for the untuned 8B teacher, and announces 26.2\% of the risks in the next 10 to 30 seconds before they appear. Of the generated advisories, 97.7\% conform to the provincial traffic rules and guidance. These findings suggest that the framework can produce rule-conforming training data for compact driving advisory models. 

[This manuscript has been submitted to the 38th IEEE Intelligent Vehicles Symposium (IV 2027).]
\end{abstract}

\section{Introduction}
Long-haul truck drivers often operate for extended periods under changing traffic, weather, and lighting conditions. Given the longer stopping distances and higher centers of gravity of heavy-duty trucks compared with passenger vehicles, timely and reliable safety advisories are essential. An intelligent co-driver system could support drivers with concise warnings about hazards such as a rapidly closing gap, a slippery road, or a vehicle stopped on the shoulder. Such a system must also anticipate hazards before they are fully visible and follow jurisdiction-specific traffic regulations.

Recent vision-language models (VLMs) have shown promising driving-scene understanding and decision-making on question-answering benchmarks~\cite{qian_nuscenesqa_2024,marcu_lingoqa_2024,sima_drivelm_2024}. However, their outputs may contain ungrounded scene interpretations~\cite{bai_hallucination_2025} or recommend actions that conflict with driving conditions and regulations. 
These benchmarks primarily assess scene understanding and reasoning rather than the validity of recommended actions, while regulation-aware approaches such as DriveReg~\cite{cai_driving_2024} query a large language model (LLM) at every decision step and reason solely over the current scene. A compact onboard model cannot afford such queries and must instead learn from training data whose advisories already incorporate the applicable rules and impending hazards.

Such training data requires structured driving evidence. Relevant information includes lane markings, surrounding traffic, predicted hazards, vehicle location, and applicable traffic rules~\cite{ma_geodrivebench_2026}. For long-haul trucking, certain restrictions on lane use and overtaking may vary with location and road type. However, existing datasets provide limited support for jointly representing these elements.
Public driving datasets with lane-marking annotations or language-based supervision primarily focus on passenger vehicles~\cite{gamerdinger_datasets_2026,sima_drivelm_2024,marcu_lingoqa_2024}, while TruckScenes~\cite{fent_truckscenes_2024} lacks the lane-marking type and language annotations required for this application.

In the proposed framework, an 8B vision-language teacher writes driving decisions and spoken advisories from the scene labels, the risk forecast, and the retrieved provincial passages, cites the passages that apply, and keeps only the records that pass a consistency check. A compact 3B-parameter student model is then distilled from these records to run on the vehicle.
The framework addresses two research questions (RQs): (i) How can structured driving evidence, risk forecasts, and applicable traffic rules be used to construct training data for truck driving advisories? (ii) Are the generated advisories consistent with applicable traffic rules and guidance?

The contributions of this study are summarized as follows:
\begin{itemize}
    \item A data generation framework that turns dashcam video into checked, rule-grounded training records for truck driving advisories.
    \item A dashcam dataset from four Canadian provinces, with manually annotated lane markings for the lane detector and driving segments labeled by the framework.
    \item A lane detector that assigns lane-marking types from frozen DINOv2 features.
    \item A compact 3B-parameter student model distilled from these records, evaluated on held-out driving data from all four provinces.
\end{itemize}


\section{Dataset}
\label{sec:dataset}
\begin{figure}[ht]
\centering
\includegraphics[width=\columnwidth]{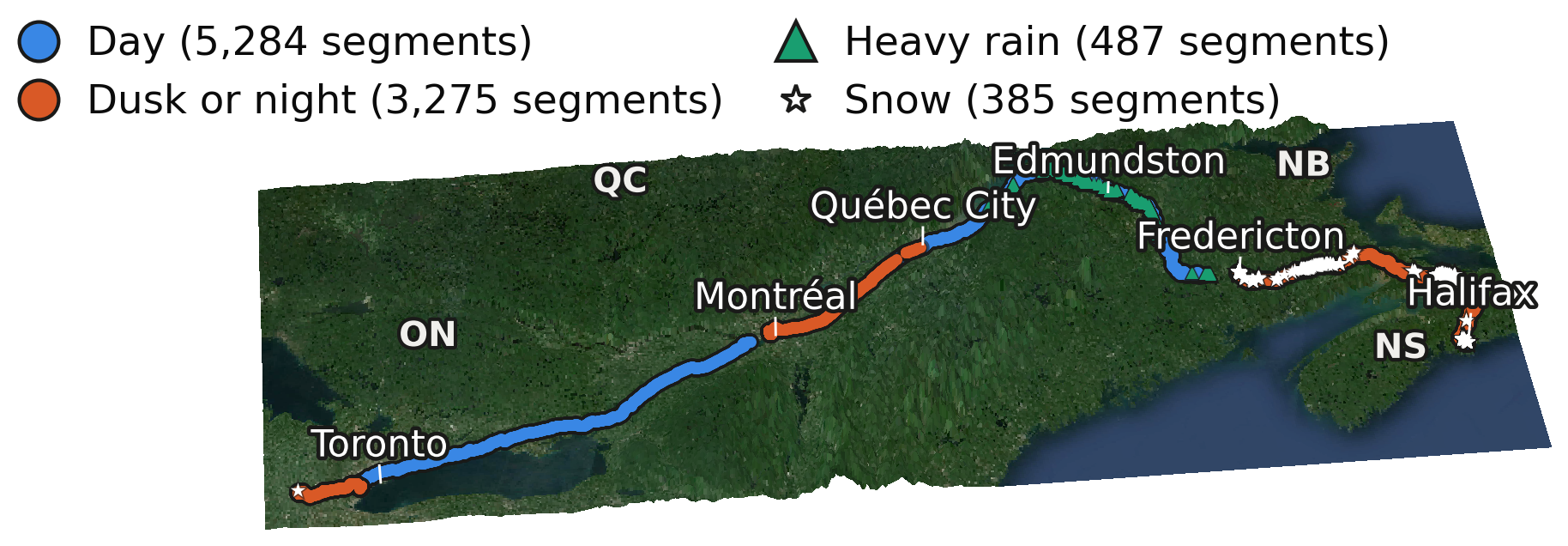}
\caption{Distribution of driving segments with valid GPS data across lighting and weather conditions.}
\label{fig:map}
\end{figure}

Both subsets, lane markings and driving advisories, were recorded with a forward-facing dashcam along a freight route through four Canadian provinces: Nova Scotia (NS), New Brunswick (NB), Quebec (QC), and Ontario (ON), as illustrated in Fig.~\ref{fig:map}. 
Table~\ref{tab:datasets} compares the dataset with public driving datasets in lane-marking type annotations (Type), question answering and driving advisories (QA), and jurisdiction-specific traffic rules (Rules).

\begin{table}[ht]
\centering
\caption{Comparison with existing driving datasets.}
\label{tab:datasets}
\footnotesize
\setlength{\tabcolsep}{3.2pt}
\begin{tabular}{@{}lcccc@{}}
\toprule
Dataset & Region & Type & QA & Rules \\
\midrule
TuSimple~\cite{tusimple_benchmark_2017} & US & \xmark & \xmark & \xmark \\
BDD100K~\cite{yu_bdd100k_2020} & US & \cmark & \xmark & \xmark \\
OpenLane~\cite{chen_persformer_2022} & US & \cmark & \xmark & \xmark \\
CADC~\cite{pitropov_cadc_2021} & CA & \xmark & \xmark & \xmark \\
Boreas~\cite{burnett_boreas_2023} & CA & \xmark & \xmark & \xmark \\
TruckScenes~\cite{fent_truckscenes_2024} & EU & \xmark & \xmark & \xmark \\
NuScenes-QA~\cite{qian_nuscenesqa_2024} & US, SG & \xmark & \cmark & \xmark \\
DriveLM~\cite{sima_drivelm_2024} & US, SG & \xmark & \cmark & \xmark \\
LingoQA~\cite{marcu_lingoqa_2024} & UK & \xmark & \cmark & \xmark \\
GeoDrive-Bench~\cite{ma_geodrivebench_2026} & 6 countries & \xmark & \cmark & country \\
\midrule
Proposed & CA & \cmark & \cmark & province \\
\bottomrule
\end{tabular}
\end{table}

The \textit{lane-marking} subset comprises 11,142 frames at a resolution of $2592\times1944$ pixels. Each lane marking is annotated as a polyline in the image with one of five types, solid white (14,754 instances), dashed white (12,822), solid yellow (10,556), double yellow (2,056), and high-occupancy vehicle (HOV) lane markings (1,997). To avoid leakage between adjacent frames, it is split into clips (a continuous video excerpt), and 41 clips with 1,833 frames are held out for evaluation.

The \textit{driving-advisory} subset comprises 1,623 one-minute videos from 35 trips, i.e., continuous drives, cut into 9,494 driving segments of 10 seconds each. These include 2,637 segments from NB, 1,827 from NS, 2,605 from QC, and 2,425 from ON. Each segment takes the province of its GPS position. Data from NB and NS were collected primarily in winter, whereas those from QC and ON were recorded in spring. Because season and province are thus correlated, the test set is drawn from every trip (Section~\ref{sec:exp}).

\begin{figure*}[ht]
\centering
\includegraphics[width=\textwidth]{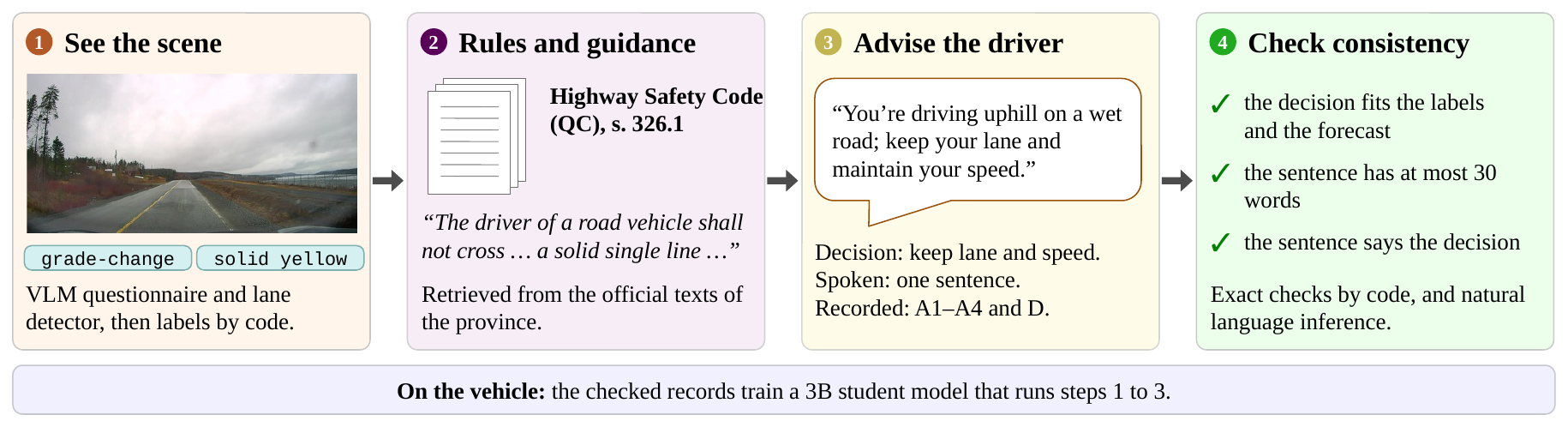}
\caption{Overview of the proposed framework.}
\label{fig:overview}
\end{figure*}

\begin{figure*}[t]
\centering
\includegraphics[width=\textwidth]{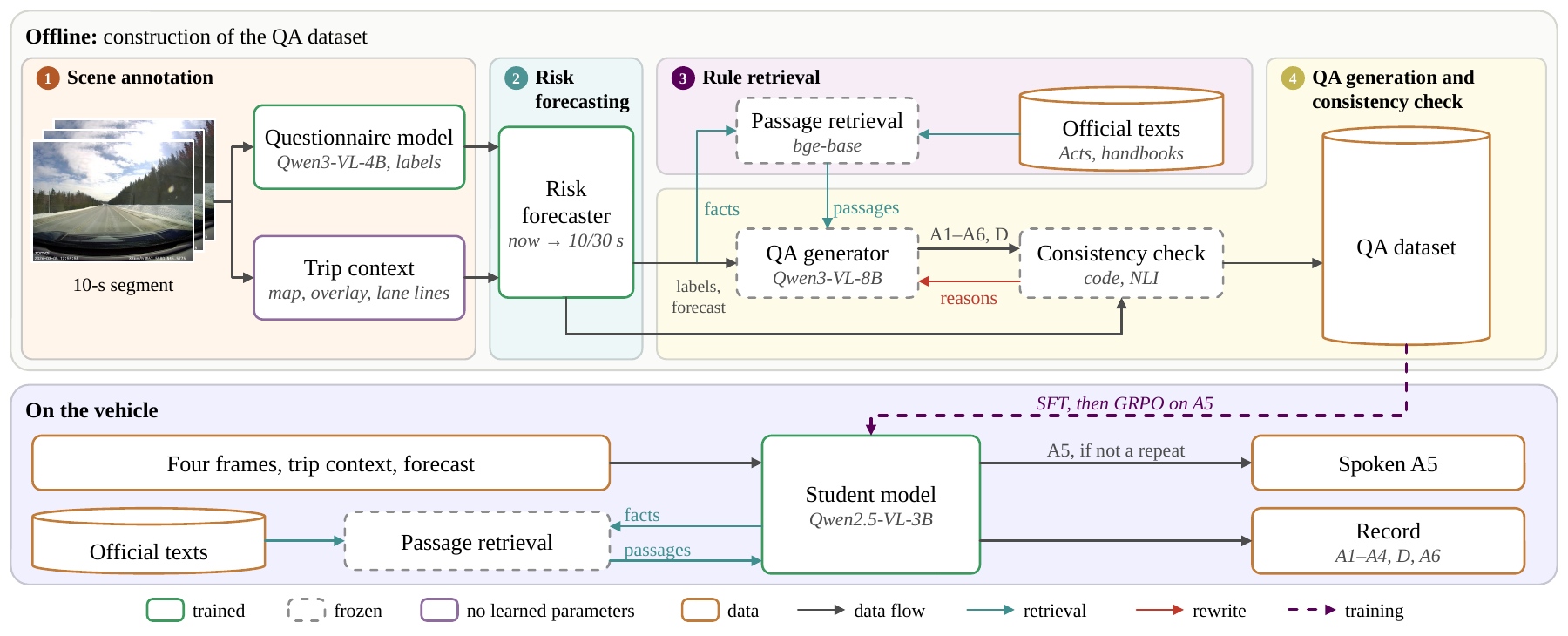}
\caption{Construction of the QA dataset and the student model. Facts are the detected labels and lane-line types used as retrieval queries, and the overlay is the GPS and speed text on the video.}
\label{fig:pipeline}
\end{figure*}

\section{Method}
\label{sec:method}
Fig.~\ref{fig:overview} gives an overview with an example and Fig.~\ref{fig:pipeline} details the framework. Recorded videos are converted offline into a question-answering (QA) dataset (Sections~\ref{sec:perception} to \ref{sec:qagen}), from which a student model is distilled (Section~\ref{sec:student}).

\subsection{Lane Detector}
\label{sec:perception}
The lane detector uses a DINOv2 ViT-L/14 backbone with registers~\cite{oquab_dinov2_2024}, kept frozen to preserve its pretrained representations. Its patch features are reduced to 128 dimensions by principal component analysis (PCA) and passed to a light head with three dilated convolutional layers and a layer of $1\times9$ and $9\times1$ convolutions that aggregates features along the rows and columns of the patch grid. For each patch, the head predicts a confidence, the offset to the bottom end of the marking, and an embedding that separates markings. A non-learned decoder then groups patches based on their embeddings and fits a smoothed curve to each detected lane marking. Each curve takes the type whose prototypes, found by $k$-means over the features of annotated training lines, best match its features:
\begin{equation}
\hat{t} = \arg\max_{t}\ \frac{1}{M}\sum_{m=1}^{M} \max_{j}\ \big\langle \mathbf{f}_m, \mathbf{p}^{t}_j \big\rangle
\label{eq:type}
\end{equation}
where $\mathbf{f}_m$ denotes the feature vector of the $m$-th patch among the $M$ patches along the curve, $\mathbf{p}^{t}_j$ is the $j$-th prototype of type $t$, and $\langle\cdot,\cdot\rangle$ represents cosine similarity.
A new lane-marking type, such as HOV, can therefore be incorporated by adding its prototypes, without retraining the detector.

\subsection{Questionnaire, Labels, and Trip Context}
\label{sec:annotation}
Scene annotation has four steps. First, the questionnaire model, Qwen3-VL-4B~\cite{bai_qwen3vl_2025}, answers 19 forced-choice questions covering road layout, surrounding vehicles, road surface, roadside environment, weather, and lighting. The questionnaire model is fine-tuned with LoRA~\cite{hu_lora_2022} on MAN TruckScenes~\cite{fent_truckscenes_2024}, whose 3D bounding boxes and scene tags provide ground-truth supervision for a subset of the questions. Second, fixed rules derive 27 scene labels from the answers. Third, GPS coordinates, speed, and timestamps are extracted from the video overlay, and an offline map resolves the province and posted speed limit. Fourth, the lane detector identifies the ego-lane marking types. Together, these outputs form a structured trip context, such as \texttt{[QC | speed limit 100 | current 111 | May/spring | left line solid yellow | right line solid white]}.

\subsection{Risk Forecaster}
\label{sec:wm}

A risk forecaster estimates the probability that a currently absent risk label will emerge within the next 10\,s or 30\,s. Each segment $t$ is represented by a vector $\mathbf{x}_t = [\mathbf{b}_t; \mathbf{c}_t]$ of indicators of the labels $\mathbf{b}_t$ and the trip context $\mathbf{c}_t$. For a label $l$ and a window of $w \in \{1, 3\}$ segments, a calibrated logistic regression on the standardized vectors $\hat{\mathbf{x}}_t$ of the last six segments gives
\begin{equation}
\hat{p}^{(w)}_{t,l} = \sigma\big(\alpha_{w,l}(\mathbf{v}_{w,l}^{\top}\hat{\mathbf{x}}_t + c_{w,l}) + \beta_{w,l}\big)
\label{eq:forecast}
\end{equation}
where $\sigma$ is the logistic function, $\mathbf{v}_{w,l}$ and $c_{w,l}$ are trained with the binary cross-entropy, and $\alpha_{w,l}$ and $\beta_{w,l}$ are fitted on held-out trips by Platt scaling. A risk is rated likely when $\hat{p} \geq 0.7$ and possible when $\hat{p} \geq 0.4$, in both cases at no less than twice its base rate in the training trips, as in \texttt{[forecast: close-following possible within 30 s]}. When no risk is rated, the forecast is \texttt{[forecast: calm]}. The forecasts in the records come from five models, each trained on four fifths of the trips and applied to the rest, so no segment is forecast by a model trained on its own trip.

\subsection{QA Generation and Consistency Check}
\label{sec:qagen}

Taking four video frames, scene labels, trip context, and the risk forecast as input, the QA generator, an 8B teacher (Qwen3-VL-8B~\cite{bai_qwen3vl_2025} used without fine-tuning), produces the multi-turn responses structured in Table~\ref{tab:fiveqa}, following the reasoning chain of DriveLM~\cite{sima_drivelm_2024} from scene description to advisory. For rule retrieval, official traffic acts and driver handbooks are chunked into passages and embedded using the pretrained bge-base-en-v1.5~\cite{xiao_cpack_2024}. Each detected fact, such as a solid or HOV line or a risk label, is a separate query that retrieves the three most relevant provincial passages. Code writes A2 from the labels and A6 from the cited passages, and the teacher writes the other answers and the decision D. Fixed labels trigger each risk category of A3, for example close following triggers spatial management. The teacher also reads the labels of the preceding segment for temporal consistency, and it samples four candidate answer sets per segment.

A consistency check then tests each candidate: A3 must follow from the labels in A2 and the forecast, D must not take an action that the forecast forbids, every cited passage must be among the retrieved passages, and A5 must state D, as judged by a pretrained natural language inference (NLI) model, add no other action, and stay within 30 words. A likely forecast risk forbids the action leading into it. When no candidate passes, the teacher rewrites the answers once with the reasons, and a segment that still fails is discarded.

\begin{table}[t]
\centering
\caption{Questions and answers of a record.}
\label{tab:fiveqa}
\footnotesize
\renewcommand{\arraystretch}{1.15}
\setlength{\tabcolsep}{3pt}
\begin{tabular}{@{}l>{\raggedright\arraybackslash}p{3.3cm}>{\raggedright\arraybackslash}p{4.3cm}@{}}
\toprule
 & Question & Answer schema \\
\midrule
A1 & What is the current road condition? & one sentence \\
A2 & Which labels apply? & labels, each with the questionnaire answer that set it (by code) \\
A3 & What is the primary risk, and how severe is it? & one of eight risk categories or no risk, severity 0 (none) to 3 (critical) \\
A4 & Why? & observation and inference \\
D & -- & lateral action (keep, left, right), longitudinal action (maintain, slow), cited passages \\
A5 & What should the driver do? & one spoken sentence of at most 30 words \\
A6 & Which official passages apply? & text of the cited passages (by code) \\
\bottomrule
\end{tabular}
\end{table}

\subsection{Student Model}
\label{sec:student}

To reduce the model size onboard, the records that pass the check are distilled into a student model, Qwen2.5-VL-3B~\cite{bai_qwen25vl_2025}. The student reads four frames of the segment, the trip context, and the forecast, and writes A1 and A2. The passages retrieved for the facts of its A2 and the trip context are then added to the conversation, and the student writes A3, A4, D, and A5. The student is trained with LoRA on both of these turns by supervised fine-tuning (SFT),
\begin{equation}
\mathcal{L}_{\mathrm{SFT}} = -\sum_{i=1}^{T} \log \pi_\theta\big(y_i \mid x, y_{<i}\big)
\end{equation}
where $x$ is the input and $y = (y_1, \ldots, y_T)$ are the tokens of the reference answers.

However, SFT also inherits the verbose, generic phrasing of the 8B teacher. To address this, a second training stage optimizes A5 via group relative policy optimization (GRPO)~\cite{shao_deepseekmath_2024}. During GRPO, the preceding outputs (A1--A4 and D) are fixed to the greedy SFT prediction, denoted collectively as $a_{1:4}$, while $G=4$ candidate advisories $\{o_i\}_{i=1}^G$ are sampled to complete the response. Each candidate is weighted by its group-standardized advantage $\hat{A}_i$, and an auxiliary SFT loss over $a_{1:4}$ with weight $\lambda_{\mathrm{a}} = 1$ is included to prevent drift in the structured scene assessment:
\begin{equation}
\begin{aligned}
\mathcal{L}_{\mathrm{GRPO}} = {} & -\frac{1}{G}\sum_{i=1}^{G} \frac{\hat A_i}{|o_i|} \sum_{t} \log \pi_\theta(o_{i,t} \mid x, a_{1:4}, o_{i,<t}) \\
 & - \frac{\lambda_{\mathrm{a}}}{|a_{1:4}|}\sum_{t} \log \pi_\theta(a_t \mid x, a_{<t})
\end{aligned}
\end{equation}

Qwen3-8B~\cite{yang_qwen3_2025} ranks each group in two opposite orders by how natural the advisories sound when spoken and how readily a driver would accept them, preferring the shorter of two equally good ones. An advisory at mean relative rank $q$ (0 best, 1 worst) receives the reward $1-q$, and one that states a speed value or names another province receives $-1$. On the vehicle, the GRPO-trained LoRA adapter writes A5, which is spoken unless it repeats the most recent advisory, and all answers are recorded.

\section{Experiments}
\label{sec:exp}



For the student model, the contiguous middle 20\% of segments in every trip (1,936) is held out for testing, and the remaining 7,046 segments that pass the consistency check are used for training. The risk forecaster is tested on the latest trips (30\% of the segments) and validated on the preceding 15\%. The questionnaire model is fine-tuned on 1,764 clips from 61 TruckScenes~\cite{fent_truckscenes_2024} drives and evaluated on 314 clips from 20 held-out drives.
All metrics except conformity and the comparison of wording are computed programmatically. The pass rate is the share of outputs that pass the consistency check with the labels the model itself writes in A2. The miss rate is the share of risk-bearing segments with a missed risk or an unparseable output, and the false-alarm rate is the share of risk-free segments that receive a warning; whether a segment bears a risk, and which categories count as correct, follows by code from its questionnaire labels, or from the forecast when no label triggers a category. The anticipation rate is the share of risks emerging within the next 10 to 30\,s that the advisory warns of in advance. Conformity is manually assessed on 300 test segments involving applicable traffic rules (75 per province), 100 per annotator. They are chosen from the scene alone, and an advisory conforms unless it conflicts with an applicable rule or guidance or omits a clearly required action; a speed above the limit or a wet road alone does not count. All experiments run on a single NVIDIA RTX 5080 GPU with 16~GB of memory.

\begin{figure*}[ht]
\centering
\includegraphics[width=\textwidth]{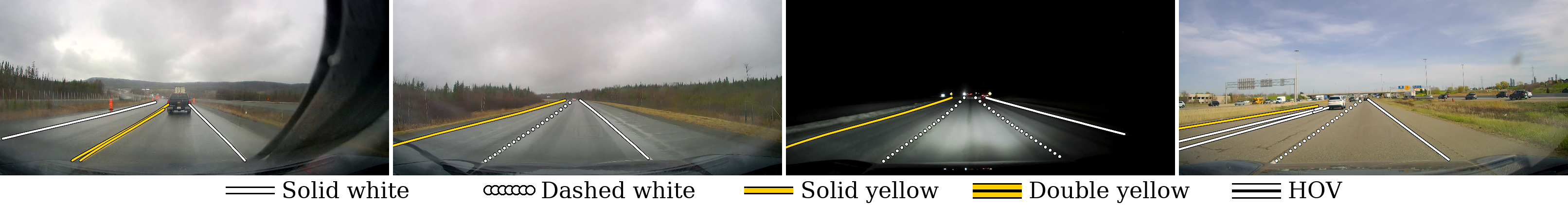}
\caption{Lane lines and their predicted types on held-out clips.}
\label{fig:lane}
\end{figure*}

\subsection{Lane Detector}
Table~\ref{tab:lane} compares the proposed lane detector with CLRerNet~\cite{honda_clrernet_2024}, with the same training routine and the TuSimple metric~\cite{tusimple_benchmark_2017}. Fig.~\ref{fig:lane} shows detections on held-out clips in rain, at night, and with an HOV lane. Our detector leads by 5.66 points on the proposed dataset and by 2.7 points when both use a quarter of the training clips (88.4\% against 85.7\%). In zero-shot transfer between the two datasets, it leads by 22.42 and 17.81 points, while CLRerNet is ahead by 1.04 points on TuSimple. Without a trained classifier, the type prototypes classify 95.02\% of the 5,906 annotated lane lines in the held-out clips correctly, and the HOV prototypes, built from two training clips, reach a recall of 100\% at a precision of 91.5\%. The margin is largest in zero-shot transfer, so the frozen features depend less on the appearance of the training roads than the end-to-end trained CLRerNet, which runs about 100 times faster; our detector is still fast enough for one advisory per 10-second segment (Section~IV-E).

\begin{table}[ht]
\centering
\caption{Lane detection accuracy (\%) and process speed.}
\label{tab:lane}
\footnotesize
\renewcommand{\arraystretch}{1.15}
\begin{tabular}{@{}lcc@{}}
\toprule
Train $\rightarrow$ test & Proposed & CLRerNet~\cite{honda_clrernet_2024} \\
\midrule
Proposed $\rightarrow$ proposed & 92.44 & 86.78 \\
Proposed $\rightarrow$ TuSimple, zero-shot & 75.19 & 52.77 \\
TuSimple $\rightarrow$ proposed, zero-shot & 82.81 & 65.00 \\
TuSimple $\rightarrow$ TuSimple & 95.13 & 96.17 \\
\midrule
Images per second & 1.59 & 161 \\
\bottomrule
\end{tabular}
\end{table}

\subsection{Annotation Pipeline}

\textit{Questionnaire.} On the held-out drives of TruckScenes, fine-tuning raises the mean accuracy of the questionnaire model from 0.693 to 0.811 and lowers the false-alarm rate of \texttt{close-following} from 18.2\% to 8.2\%. Compared to a free-form description by the same model, classified into labels by Qwen3-8B~\cite{yang_qwen3_2025}, the forced-choice questionnaire achieves a mean recall of 63.1\% versus 53.6\%, at a false-alarm rate of 10.2\% versus 14.1\%. The forced-choice questionnaire gains most on hazards that a free description tends to leave unmentioned, such as roadside activity (recall 53.8\% against 7.7\%) and a closing gap (55.6\% against 0\%).

\textit{QA dataset.} During QA generation, 90.7\% of the segments pass the consistency check on the first attempt, 2.4\% pass after a single rewrite, and 6.9\% are discarded. The teacher cites at least one passage in 26.1\% of the retained records. The rewrite recovers about a quarter of the failing segments.

\textit{Risk forecaster.} The onset AUC, which measures how well the forecast ranks segments in which a label absent from the previous segment appears within the window, averages 0.735 over labels and windows on the test trips, against 0.703 for the frequency of each label over the preceding five segments. The forecast is most reliable for lead braking (onset AUC 0.88 within 30\,s), roadside activity (0.84), and glare (0.80), and least reliable for close following (0.63) and lane closures (0.60), which depend on the momentary behavior of other vehicles or on work zones that are not yet visible.

\subsection{Student Model}
The two untuned models in Table~\ref{tab:main} receive the same input as the student together with the label and category vocabulary. The 3B model rarely gives a valid category or decision, so it misses 98.1\% of the risks, and its Viol. is 0.0 only because its decisions can seldom be parsed. The 8B teacher passes the consistency check in only 20.2\% of the segments. After SFT, 94.6\% of the student's outputs pass the check (Pass), and 74.4\% name a category that the labels or forecast trigger (Cat.). The student misses 15.4\% of the risks and warns in 27.7\% of the segments without risk (FA). It is better than the untuned teacher on every metric, so the checked records, not the model size, make the advisories consistent.

In a visual check of 30 random false alarms, 29 warn about a vehicle that is far ahead or in the adjacent lane. In 103 of the 153 false alarms, D keeps the speed and the lane, so the advisory only points out the vehicle and asks for no action. The anticipation rate (Ant.) is 26.2\%. In 1.4\% of the segments, D takes an action that the forecast forbids (Viol.), and no advisory refers to another province. D cites a passage in 24.3\% of the segments, always one that was retrieved.

GRPO leaves the other metrics at the level of SFT and raises the pass rate to 95.6\%. Among the 1,530 test segments in which the two advisories differ, Qwen3-8B, judging each pair in both orders, prefers the GRPO advisory in 824 and the SFT advisory in 121, with 585 ties, at the same mean length of 11 words. Together with Section~IV-B, these results answer RQ (i), since checked records built from scene labels, risk forecasts, and retrieved rules are enough to train a 3B model that gives consistent advisories.

\begin{table}[ht]
\centering
\caption{Models on the held-out segments of all trips (\%).}
\label{tab:main}
\footnotesize
\setlength{\tabcolsep}{3.4pt}
\begin{tabular}{@{}lcccccc@{}}
\toprule
Model & Pass & Cat. & Miss & FA & Ant. & Viol. \\
\midrule
Qwen2.5-VL-3B, untuned & 0.0 & 0.5 & 98.1 & 1.1 & 8.1 & 0.0 \\
Qwen3-VL-8B teacher, untuned & 20.2 & 40.8 & 55.4 & 30.2 & 19.0 & 1.9 \\
\midrule
Student model (SFT) & 94.6 & 74.4 & 15.4 & 27.7 & 26.2 & 1.4 \\
\quad with GRPO on the wording & 95.6 & 74.5 & 15.2 & 27.7 & 26.1 & 1.4 \\
\bottomrule
\end{tabular}
\end{table}

\begin{table*}[t]
\centering
\caption{Case study on four test segments. Rules lists the passages that D cites, and Retrieved marks passages retrieved but not cited.}
\label{tab:cases}
\scriptsize
\resizebox{\textwidth}{!}{%
\setlength{\tabcolsep}{3pt}
\begin{tabular}{
  >{\raggedright\arraybackslash}m{1.5cm}
  >{\raggedright\arraybackslash}m{0.215\linewidth}
  >{\raggedright\arraybackslash}m{0.215\linewidth}
  >{\raggedright\arraybackslash}m{0.215\linewidth}
  >{\raggedright\arraybackslash}m{0.215\linewidth}}
\toprule
 & \textbf{Braking Lead at Night (QC)} & \textbf{Forecast Close Following (ON)} & \textbf{Forecast Braking (NB)} & \textbf{Curve Above the Limit (NS)} \\
\midrule
\textbf{Dashcam Frames} &
    \includegraphics[width=\linewidth]{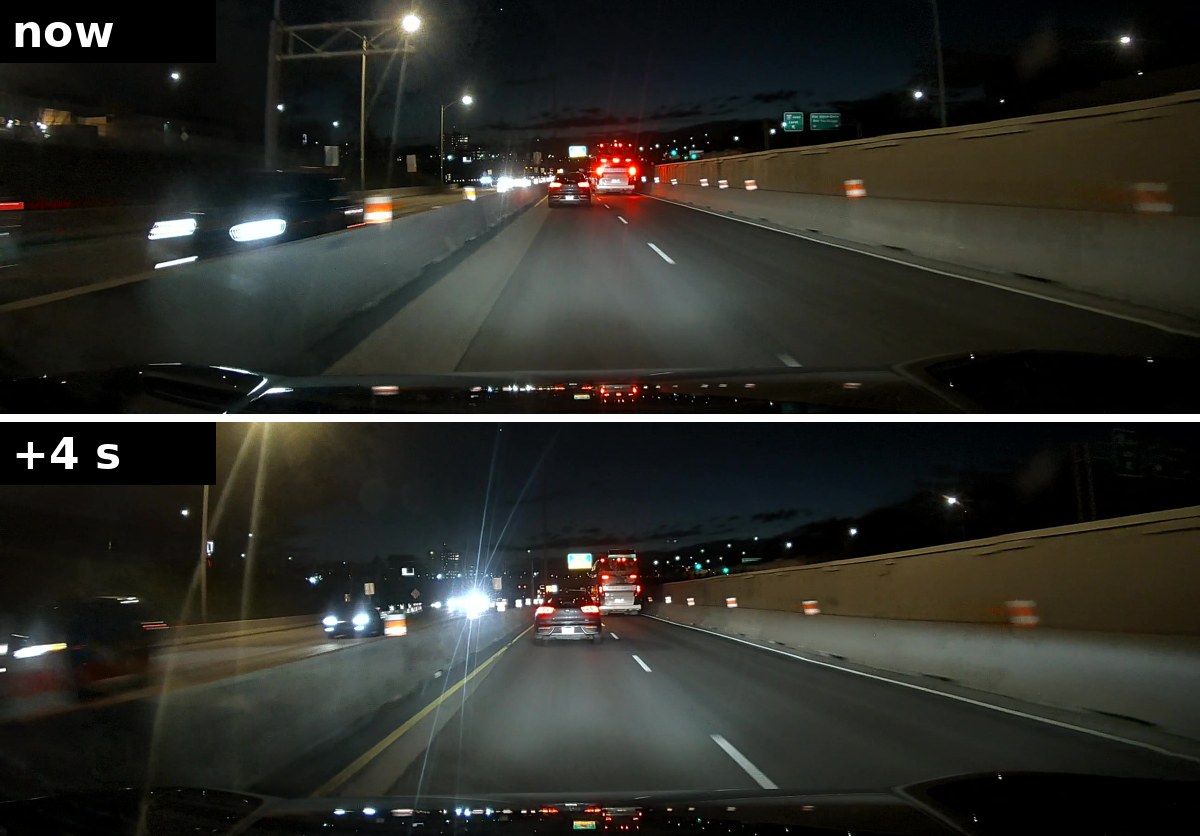} &
    \includegraphics[width=\linewidth]{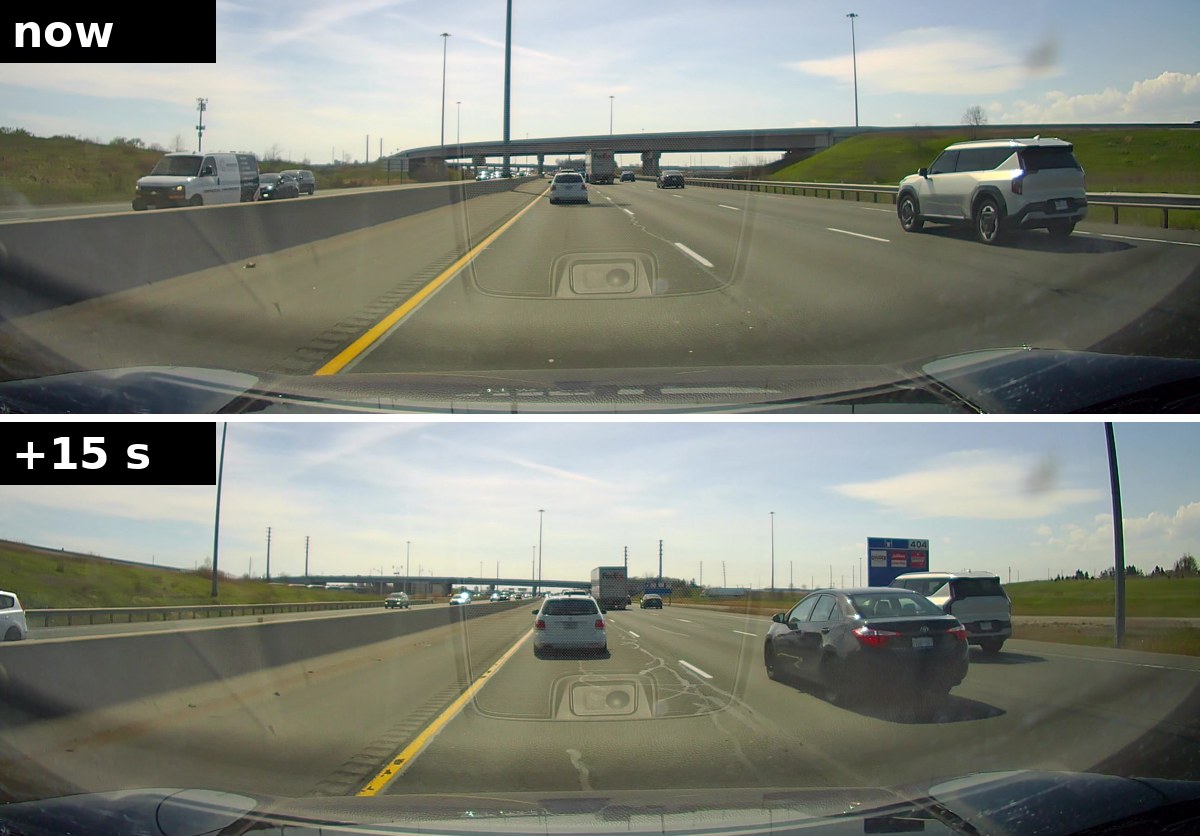} &
    \includegraphics[width=\linewidth]{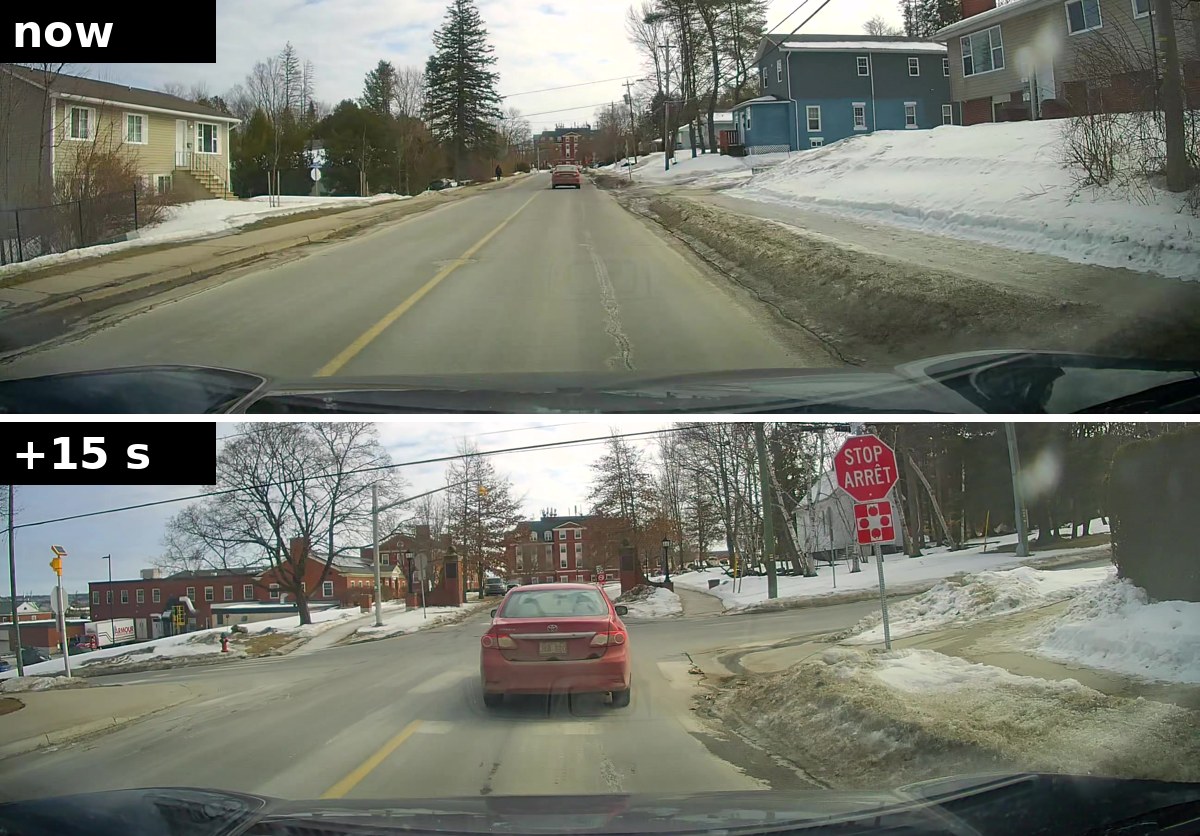} &
    \includegraphics[width=\linewidth]{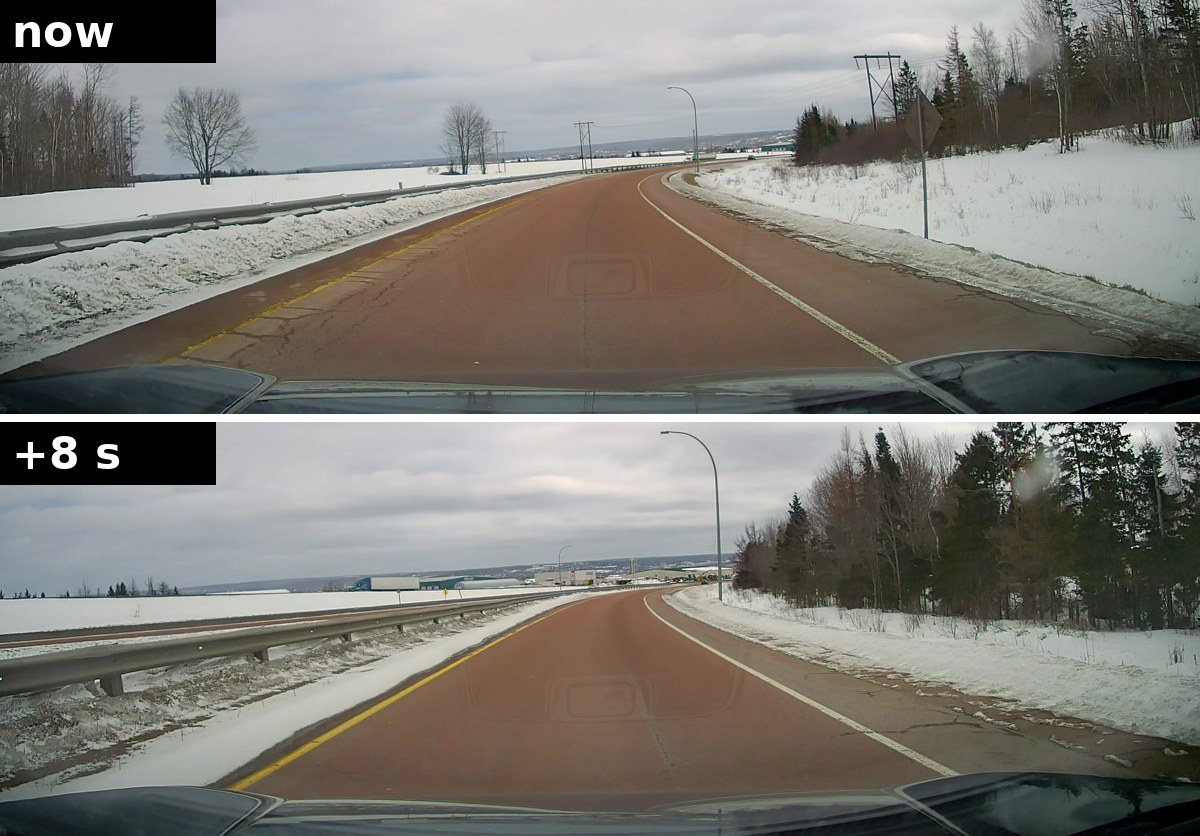} \\
\midrule
\textbf{Trip Context} &
  {\tiny\texttt{[QC | speed limit 70 | current 65 | May/spring | left line solid yellow | right line dashed white]}}\newline {\tiny\texttt{[forecast: calm]}} &
  {\tiny\texttt{[ON | speed limit 110 | current 99 | May/spring | left line solid yellow | right line dashed white]}}\newline {\tiny\texttt{[forecast: close-following possible within 30 s]}} &
  {\tiny\texttt{[NB | speed limit 50 | current 47 | March/winter | left line solid yellow | right line solid white]}}\newline {\tiny\texttt{[forecast: close-following possible within 10 s · lead-braking possible within 10 s · lead-lateral-instability possible within 30 s]}} &
  {\tiny\texttt{[NS | speed limit 90 | current 99 | March/winter | left line solid yellow | right line solid white | curve ahead]}}\newline {\tiny\texttt{[forecast: calm]}} \\
\midrule
\textbf{Labels and\newline Reasoning} &
  \textbf{A2:} highway, close-following, lead-braking, lead-approaching, lead-lateral-instability, night\newline \textbf{A3:} spatial management, severity 2\newline Lead vehicle is braking and drifting laterally while being closely followed, requiring caution. &
  \textbf{A2:} highway, lead-approaching\newline \textbf{A3:} spatial management, severity 2\newline The lead vehicle is ahead and the gap is closing, requiring caution due to the approaching distance. &
  \textbf{A2:} urban, lead-approaching, grade-change\newline \textbf{A3:} spatial management, severity 2\newline Lead vehicle ahead with brake lights off, gap closing, and forecast indicates close-following possible within 10 seconds. &
  \textbf{A2:} rural, curve\newline \textbf{A3:} curve speed, severity 1\newline Curve ahead as indicated by map and road alignment. Speed is high for the curve. Reduce speed to maintain control. \\
\midrule
\textbf{Rules} &
  QC Highway Safety Code s.~335 (following distance)\newline Retrieved: s.~326.1 (no crossing a solid line) &
  -- &
  Retrieved: NB Driver's Handbook, p.~5 (no passing when the solid line is on your side) &
  NS Driver's Handbook, pp.~68--69 (speed in curves)\newline Retrieved: p.~102 (solid yellow line at the left edge of a divided highway) \\
\midrule
\textbf{Safety\newline Advisory} &
  The vehicle ahead is braking and drifting; ease off to maintain distance. &
  The vehicle ahead is closing the gap; ease off to increase your following distance. &
  Vehicle ahead is closing the gap; ease off to increase following distance. &
  Curve ahead, slow down; keep your lane. \\
\bottomrule
\end{tabular}
}
\end{table*}

\subsection{Conformity}

Of the 300 advisories, 97.7\% conform to the provincial rules and guidance, with 98.7\% in New Brunswick, 98.7\% in Nova Scotia, 96.0\% in Quebec, and 97.3\% in Ontario. A further 32 advisories, eight per province, are assessed by all three annotators, who agree on all of them. The seven nonconforming advisories keep the speed although A2 names a work zone (four), a sharp curve (one), or a close or cutting-in lead vehicle (two). The consistency check misses them because it forbids actions only on the basis of the forecast. The generated advisories thus largely conform to the provincial rules and guidance, which answers RQ (ii).

\subsection{Qualitative Results and Latency}

Table~\ref{tab:cases} shows one test segment from each province. The student cites the Quebec following-distance rule and the Nova Scotia guidance on curve speeds. For the solid yellow line reported by the detector on the left, retrieval returns the passage of each province, which forbids crossing or passing in Quebec and New Brunswick and marks the left edge of a divided highway in Nova Scotia. In the Ontario and New Brunswick segments, close following is not yet labeled, though the forecast rates it as possible. The student model proactively eases off, and close following indeed occurs in the next segment, where traffic slows from 99 to 77\,km/h in Ontario and the lead vehicle stops at a stop sign in New Brunswick. On an RTX 5080, a segment takes 2.5\,s for the lane detector, 1.2\,s for the student model, and 0.24\,s for the A5 pass of GRPO, against 2.6\,s for the 8B teacher alone.

\section{Conclusion}
This paper presented a verification-guided data generation and distillation framework for rule-grounded truck driving advisories, supported by a dashcam dataset collected across four Canadian provinces. The central contribution is the integration of structured driving evidence, risk forecasts, and applicable traffic rules into a verification process that filters candidate advisories before they are used to train a compact student model. Experimental results demonstrate that the distilled model can generate advisories with high consistency under the verification criteria, while sampled recommendations show strong conformity with provincial rules and guidance.
Future work will expand the dataset across seasons and driving conditions, and evaluate the framework through on-road testing.

\bibliographystyle{IEEEtran}
\bibliography{references}

\end{document}